\documentclass[runningheads]{llncs}
\usepackage{graphicx}
\usepackage{comment}
\usepackage{amsmath,amssymb}
\usepackage{color}

\usepackage{multirow}
\usepackage{algorithmic}
\usepackage{algorithm}

\usepackage{microtype, url}

\makeatletter
\renewcommand\paragraph{\@startsection{paragraph}{4}{\z@}%
                       {-12\p@ \@plus -4\p@ \@minus -4\p@}%
                       {-0.5em \@plus -0.22em \@minus -0.1em}%
                       {\normalfont\normalsize}}
\makeatother

\begin{document}
	\pagestyle{headings}
	\mainmatter
	\def\ECCVSubNumber{1022}

	\title{Weighing Counts: Sequential Crowd Counting by Reinforcement Learning\thanks{L. Liu and H. Lu contributed equally. Z. Cao is the corresponding author. Part of this work was done when L. Liu was visiting  The University of Adelaide.
\it Accepted to Proc.\ European Conf.\ Computer Vision 2020.
	}}

	%

	%
	\titlerunning{Weighing Counts: Sequential Crowd Counting by Reinforcement Learning}
	\author{Liang Liu\inst{1} \and
	Hao Lu\inst{2} \and
	Hongwei Zou\inst{1}\and
	Haipeng Xiong\inst{1}\and\\
	Zhiguo Cao\inst{1}\and
	Chunhua Shen\inst{2}}

	\authorrunning{Liu et al.}
	\institute{
	School of %
	Aritifical Intelligence \& Automation,
	Huazhong University of Science \&  Technology, China \and
	The University of Adelaide, Australia\\
	\email{\{wings, zgcao\}@hust.edu.cn}}

	\maketitle
	\begin{abstract}

		We
		formulate
		counting as a sequential decision problem and present a novel crowd counting model
		solvable
		by deep reinforcement learning. In contrast to existing counting models that directly output count values,
		we divide one-step estimation into a sequence of much
		easier
		and more  tractable sub-decision problems.
		Such sequential decision nature corresponds exactly to a physical process in reality---scale weighing. Inspired by scale weighing, we propose a novel `counting scale' termed LibraNet where the count value is analogized by weight. By virtually placing a crowd image on one side of a scale, LibraNet (agent) sequentially learns to place appropriate weights on the other side to match the %
		crowd
		count. At each step, LibraNet chooses one weight (action) from the weight box (the pre-defined action pool) according to the current crowd image features and weights placed on the scale pan (state). LibraNet is required to learn to balance the scale according to the feedback of the needle (Q values).
	    We show that LibraNet exactly implements scale weighing by visualizing the decision process
	    how
	    LibraNet chooses actions. Extensive experiments %
	    demonstrate
	    the effectiveness of our design choices and report state-of-the-art %
	    results
	    on %
	    a few
	    crowd counting benchmarks, %
	    including
	    ShanghaiTech, UCF\_CC\_50 and UCF-QNRF. We also demonstrate good cross-dataset generalization of LibraNet. Code and models are made available at \url{https://git.io/libranet}

	\end{abstract}

	\section{Introduction}\label{sec:Introduction}

	Counting is sequential decision process by nature. Dense object counts are not inferred by humans with a simple glance~\cite{chattopadhyay2017counting}.
	Instead humans count objects in a sequential manner, with initial fast counting on apparent objects (large sizes and clear appearance) and gradually slow counting on objects that are hard to recognize (small sizes or blurred appearance).
	Such a sequential decision behavior can be modeled by a physical process in reality---scale weighing.
	In scale weighing, it is easy to choose a weight when the weights placed on the scale are far from the true weight of the stuff. When placed weights are close to the true weight, %
	small and light weights are carefully chosen until the needle indicates
	the balance.
	This process decomposes a %
	difficult
	problem into a series of much more tractable sub-problems.

	\begin{figure}[!t]
		\begin{center}
			\includegraphics[width=\linewidth]{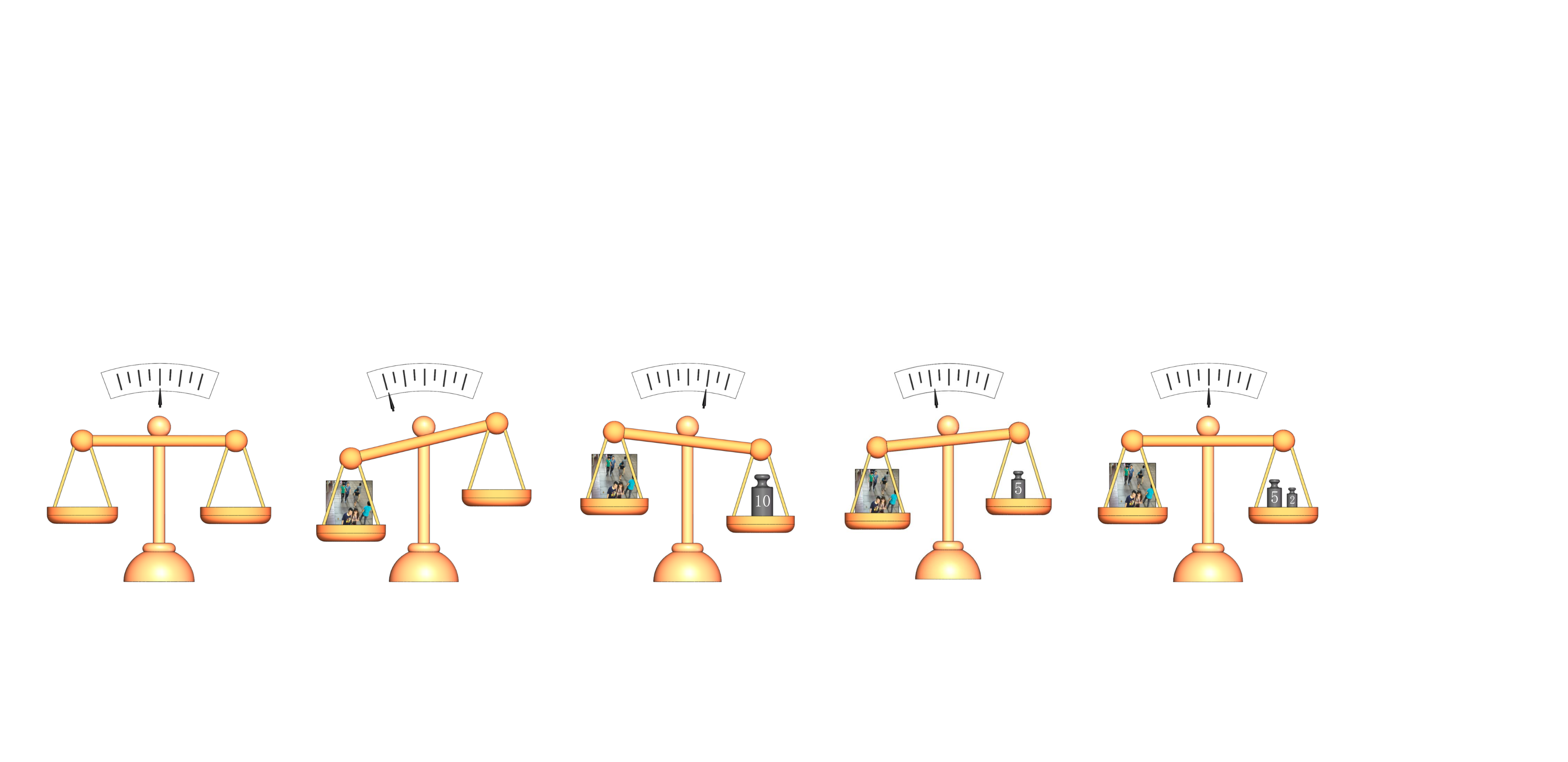}\\
		\end{center}
		\caption{
		Counting scale. We implement crowd counting as scale weighing.
		By virtually placing a crowd image (with $7$ people) on the scale, if placing a $10$ g weight on the scale pan, the scale will lean to the right; if exchanging the $10$ g weight to $5$ g, the scale instead will lean to the left.
		Finally by adding another $2$ g weight, the scale is balanced. The total weights on the scale can therefore indicate the number of crowd.}
		\label{fig:FIG1}
	\end{figure}

	Following the same spirit of human counting and scale weighing, we %
	formulate counting as a sequential decision problem and implement it as scale weighing. Indeed counting objects is like weighing stuff. In the context of crowd counting shown in Fig.~\ref{fig:FIG1}, the `stuff' is a crowd image, and the `weights' are a series of pre-defined value operators. We repeatedly choose counting `weights' to approximate the ground-truth counts until the scale is balanced. The final image count is simply a summation of placed `weights'.

	The sequential decision nature of scale weighing makes it suitable to be described by a reinforcement learning (RL) task.
	We hence propose a Deep Q-Network (DQN)~\cite{mnih2015human}-based solution, LibraNet\footnote{The naming of LibraNet is inspired by %
	the zodiac sign.
	}, to implement scale weighing and apply it to crowd counting as a `counting scale'. In particular, given a `stuff', LibraNet outputs a combination of weights step by step. In each step, a weight (action) is chosen from the weight box (the pre-defined action pool) or removed from the scale pan according to the feedback of the needle (Q values that indicate how to choose the next action). The weighing process continues until LibraNet chooses the `end' operator. The `stuff' is the image feature encoded from a crowd image, and the `end' condition meets when the summation of the weights equals/approximates to the ground-truth people count.

	We visualize how LibraNet works and illustrate that LibraNet exactly implements scale weighing. We show through extensive experiments why our choices in designing reward function
	work well, that LibraNet can be used as a plug-in to existing local counts models~\cite{xiong2019tasselnetv2,liu2019countingobject}, and that LibraNet achieves state-of-the-art performance on three crowd counting datasets, including ShanghaiTech~\cite{zhang2016single}, UCF\_CC\_50~\cite{Idrees2013Multi}, and UCF-QNRF~\cite{idrees2018composition}. We also report cross-dataset performance to verify the generalization of LibraNet.

	In summary, we show
	that
	counting can be interpreted as scale weighing and we implement scale weighing with LibraNet. To
	our knowledge, LibraNet is the first approach that
	uses RL techniques to solve crowd counting.

	\subsection{Related Work}\label{sec:Related}

	\paragraph{\textbf{Crowd Counting}.}
	Crowd counting is often tackled as a dense prediction task
	\cite{lu2019indices,lu2020index}.
	Solutions range from early attempts that detect pedestrians~\cite{Dalal2005Histograms}, regress image counts~\cite{chan2008privacy}, estimate density maps~\cite{lempitsky2010learning}, predict localized counts~\cite{chen2012feature}, to recent deep learning-based density maps estimation~\cite{li2018csrnet},
	redundant counts regression~\cite{paul2017count,lu2017tasselnet}, instance blobs localization~\cite{laradji2018blobs} and count intervals classification~\cite{liu2019countingobject,Xiong2019CLOSE}.

	Since detection typically failed on small and dense people, regression-based approaches~\cite{chan2008privacy,ryan2009crowd} were proposed. While early methods alleviated the issues of occlusion and clutter, they ignored spatial information because only the global image count was regressed. This situation eased when the concept of density map was introduced in~\cite{lempitsky2010learning}.
	Chen \textit{et al.}~\cite{chen2012feature} also introduced localized count regression by mining local feature importance and sharing visual information among spatial regions.

	With the success of Deep Convolutional Neural Networks (DCNNs), deep crowd counting models emerged. \cite{wang2015deep}	applied a CNN to crowd counting by global count regression.~\cite{zhang2015cross} presented a switchable training scheme to estimate the density map and the global count. By contrast, works of \cite{paul2017count,lu2017tasselnet} adopted redundant counting where local patches were densely sampled in a sliding-window manner during training, and the image count was obtained by normalizing redundant local counts at inference time.
	Authors of
	\cite{Liu_2019_ICCV} employed a CRF-based structured feature enhancement module and a dilated multiscale structural similarity loss to address scale variations of crowd. To alleviate perspective distortion, the work in \cite{shi2019revisiting} integrated perspective information into density regression and proposed a PACNN for efficient crowd counting. In \cite{laradji2018blobs} %
	a network is trained  to output a single blob for each person for localization.
	The work in \cite{wan2019residual}  optimized a residual signal to refine the density map.
	Instead of direct regression, authors of \cite{liu2019countingobject,Xiong2019CLOSE} reformulated it as a classification problem by discretizing local counts and classifying count intervals.

	Most existing models generate crowd counts in one step. This renders difficulties in correcting under- or over-estimated counts. Despite
	that
	there exists a method that recurrently refines density map with a spatial transformer network \cite{liuijcai2018crowd}, it does not decompose a hard task into %
	a sequence of
	easy sub-tasks and does not fully leverage the advantage of sequential counting.

	\paragraph{\textbf{Deep Reinforcement Learning}.}
	RL~\cite{diuk2008object,riedmiller2009reinforcement} is one of the %
	fundamental
	machine learning paradigms. It includes several elements, %
	namely,
	agent, environment, policy, state, action, and reward. It aims to learn policies such that an agent can receive the maximum reward when interacting with the environment. Since the work of \cite{mnih2013playing} introduced deep learning into RL,
	it has received extensive studies
	\cite{mnih2016asynchronous,mnih2015human,schulman2017proximal,wang2015dueling}.
	In particular, RL achieved %
	breakthroughs in a few areas
	such as go~\cite{silver2016mastering} and real-time strategy games~\cite{OpenAI_dota,vinyals2019grandmaster}. Recently some deep RL-based methods were also proposed to
	tackle
	computer vision %
	tasks, such as object localization~\cite{caicedo2015active}
	and
	instance segmentation~\cite{araslanov2019actor}.
	However, these RL practices in computer vision cannot be directly %
	transferred to
	crowd counting. A main reason is that there is no principled way to reformulate counting %
	into
	a sequential decision problem suitable for RL. Inspired by
	scale weighing, we fill this gap and present the first deep RL-based approach to crowd counting.
	\begin{figure*}[!t]
		\centering
		\includegraphics[width=\linewidth]{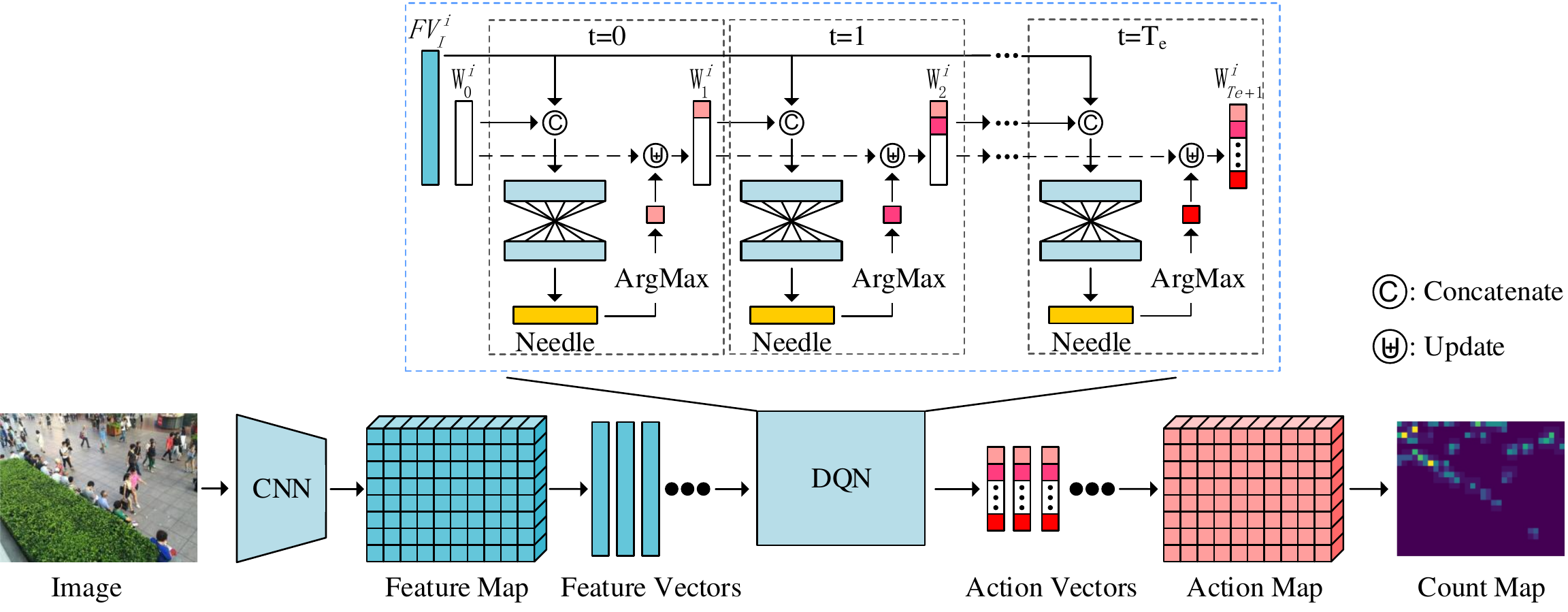}%
		\caption{Overview of LibraNet. A CNN backbone first extracts the feature map $FV_I$ of an input image $I$, then each element $FV^i_I$ of $FV_I$ is sent to a DQN. In DQN, $FV^i_I$ and a weighing vector $W_t^i$ are concatenated and sent to a $2$-layer MLP. The output of MLP is a $9$-dimensional Q value vector. We choose an action with the maximum Q value, and update $W_{t+1}^i$ per Eq.~\eqref{equ: W update}. This process repeats until the model chooses `end' or exceeds the predefined maximum step. The output action vectors can be converted to count intervals by Eq.~\eqref{equ: V_T}, and the intervals 	are further remapped to a count map	with inverse discretization~\cite{liu2019countingobject}. The image count of $I$ is acquired by summing the count map.}
		\label{fig:overview}
	\end{figure*}

	\section{Sequential Crowd Counting by Reinforcement Learning}\label{sec:Sequential Crowd Counting by Reinforcement Learning}
	Here we explain LibraNet in detail. Sec.~\ref{subsec:Sequential Counting Task Formulation} introduces the formulation of sequential counting. Sec.~\ref{subsec:Q learning} shows how to deal with this sequential task with Q-learning. Sec.~\ref{subsec:Structure} explains the network architecture, and Sec.~\ref{subsec:Implement Details} presents implementation details. An overview of our method is shown in Fig.~\ref{fig:overview}.

	\subsection{Generalized Local Count Modeling}\label{subsec:Sequential Counting Task Formulation}
	Despite that most deep counting networks treat density maps as the regression target~\cite{idrees2018composition,ma2019bayesian,sam2017switching,wang2015deep,zhang2016single}, there is another line of works pursuing the idea of local count modeling and also reporting promising results~\cite{paul2017count,liu2019countingobject,lu2017tasselnet,Xiong2019CLOSE}. LibraNet follows this local count paradigm but operates in a sequential manner. In what follows, we present a generalized perspective of local count modeling and show how we reformulate them into sequential learning.

	\paragraph{\textbf{Local Count Regression}.}
	Some previous works~\cite{paul2017count,lu2017tasselnet,xiong2019tasselnetv2} consider counting a problem of local count regression, which densely samples an image into a series of local patches then estimates the per-patch count directly. It amounts to the following optimization problem

	\begin{equation} \small \label{equ:regression}
	\underset{\theta }{\mathop{\min }}\,\sum\limits_{i\in I}{{\left|G\left(i\right)-  {N_{R}^{\theta}\left( i \right)} \right|}}\,,
	\end{equation}
	where $I$ is the input image and $i$ denotes the local patch sampled from $I$, $G\left(i\right)$ returns the ground truth count given $i$, and $N_{R}^{\theta}$ is a regression network parameterized by $\theta$.

	\paragraph{\textbf{Local Count Classification}.}
	Inspired by local count regression, counting is further formulated as a classification problem~\cite{liu2019countingobject,Xiong2019CLOSE} where local patch counts are discretized into count intervals. This process is defined by
	\begin{equation} \small \label{equ:classification}
	\underset{\theta }{\mathop{\min }}\,\sum\limits_{i\in I}{{\left| G\left(i\right)-{\rm ID}\left( {\underset{c}{\mathop{\arg \max }}\,N_{C }^{\theta}\left( i,c \right) }\right)\right|}}\,,
	\end{equation}
	where $N_{C}^{\theta}$ is a classification network parameterized by $\theta$, $c$ is the number of count intervals, and $\rm ID(\cdot)$ defines an inverse-discretization procedure that recovers the count value from the count interval~\cite{liu2019countingobject}. More details about discretization and inverse-discretization can be referred to Supplementary Materials.

	\paragraph{\textbf{Local Counting by Sequential Decision}.}
	Motivated by scale weighing, counting can be transformed into a sequential decision task. We call this a \textit{weighing task}. Instead of estimating a count value or a count interval directly, the weighing task sequentially chooses a value operation in each step from a pre-defined action pool. The sequential process terminates when the agent chooses the `ending' operation or exceeds the maximum step allowed. This task is defined by
	\begin{equation} \small\label{equ: sequential}
	\underset{\theta }{\mathop{\min }}\,\sum\limits_{i\in I}{{\left| G\left(i\right)-\sum\limits_{t=0}^{T_e}{\underset{{{a}}}{\mathop{\arg \max }}\,}N_{E }^{\theta}\left( i,{{W}_{t}^i},{{a}}\right)  \right|}}\,,
	\end{equation}
	where $N_{E }^{\theta}$ is a sequential decision network parameterized by $\theta$, ${a}$ is one of the pre-defined value operations. $T_e=\min\left(t_{m}, t_{e} \right)$ is the ending step, where $t_{m}$ is the maximum step, $t_{e}$ is the step that chooses the ending operation.
	$W_t^i$ is the weight vector that represents the chosen weights, which is initialized by a full-zero vector.
	$W_t^i$ takes the form
	\begin{equation} \small\label{equ: W update}
	{{W}_{t+1}^i}=\left\{ \begin{matrix}
	{\{0,0,0,....\}} & if \ t=0  \\
	{{W}_{t}^i}\uplus {{a}_{t}} & otherwise  \\
	\end{matrix} \right.\,,
	\end{equation}
	where ${{a}_{t}}$ is the operation chosen at the step $t$, and $\uplus$ is a weight updating operator (see also Eq.~\eqref{equ:+}). In step $T$ the count $V^{i}_T$ of the patch $i$ takes the form
	\begin{equation} \small\label{equ: V_T}
	V^{i}_T=\sum\limits_{t=0}^{T}{\underset{{{a}}}{\mathop{\arg \max }}\,}N_{E }^{\theta}\left( i,{{W}_{t}^i},{{a}}\right) =\sum\limits_{t=0}^{T}{w^i_t}\,,
	\end{equation}
	where $w^i_t$ forms $W^i_t$ such that
	\begin{equation} \small
	W^i_t=\left(w^i_0,w^i_1,...,w^i_{t-1},0,...\right)\,.
	\end{equation}
	Overall, the working flow of this weighing task is akin to scale weighing. In each step, the network $N_{E}^{\theta}$ (scale) evaluates the value difference between the image patch $i$ and the value associated with the weight vector $W_t^i$ (weights); according to the output of the network (needle), the agent chooses an action (add or remove a weight) to adjust $V^{i}_T$ to approximate the ground-truth patch count $G\left(i\right)$ until they are equal (the scale is balanced). We
	present
	more details %
	in the sequel.

	\subsection{Crowd Counting as Sequential Scale Weighing}\label{subsec:Q learning}
	We implement Eq.~\eqref{equ: sequential} within the framework of Q-leaning~\cite{mnih2015human}. The elements of Q-learning include state, action, reward and
	Q value. They correspond to the scale pan, weights, designed rewards and needle in 	scale weighing.

	\paragraph{\textbf{State (Scale Pan)}.}
	The state depicts the status of `two scale pans'---the weight vector $W_{t}^i$ and the image feature $FV^i_I$. Formally, the state $s=\{FV^i_I, W_t^i\}$.

	According to~\cite{liu2019countingobject}, the data distribution is often long-tailed in crowd counting datasets with imbalanced samples. Liu \textit{et al.}~\cite{liu2019countingobject} shows that this issue could be alleviated by quantizing local counts and treating the count intervals as the learning target. We follow this idea to check the balancing condition of the scale.

	\paragraph{\textbf{Action (Weights)}.}
	In Q-learning, an action is defined to modify the state.
	Since $FV^i_I$ is fixed in $s$ once it is extracted, the action is designed to only change $W_{t}^i$. We design an action pool in a way similar to the scale weighing system and the money system~\cite{van2001optimal}, i.e., $a=\{-10, -5, -2, -1, +1, +2, +5, +10, end\}$. It includes $8$ value operations and one ending operation (indicating the scale is balanced).
	Given a new action ${a}_{t}$, $W_t^i$ is modified by an updating operator $\uplus$
	\begin{equation}\label{equ:+}
	{W}_{t}^i\uplus {{a}_{t}}=\{w_0^i,...,w_{t-1}^i,0,0,...\}\uplus a_t=   \{w_0^i,...,w_{t-1}^i,a_t,0,...\}\,.
	\end{equation}
	$W_t^i$ records what weights are placed/removed from the scale pan before step $t-1$.

	\paragraph{\textbf{Reward Function}.}
	A reward scores the value of each action. We define two types of reward: ending reward and intermediate reward. In particular, we use a conventional \textit{ending reward} and further design three counting-specific rewards---\textit{force ending reward}, \textit{guiding reward}, and \textit{squeezing reward}.

	\paragraph{\textit{Ending Reward}.}
	Following~\cite{caicedo2015active}, we employ a conventional \textit{ending reward} to evaluate the value of the `end' action, defined by
	\begin{equation} \small \label{equ:End}
	{{R}_{e}}\left( {{E}_{t_e-1}} \right)=\left\{ \begin{matrix}
	+{{\eta }_{e}} & ~~~~~\text{if} \left| {{E}_{t_e-1}} \right|\le\epsilon_1  \\
	-{{\eta }_{e}} & \text{otherwise}  \\
	\end{matrix} \right.\,,
	\end{equation}
	where $t_e$ is the step that the agent chooses the `end' action, ${E}_{t_e-1}$ is the absolute value error between the ground-truth count $G(i)$ and the accumulated value $V^i_{t_e-1}$, and $\epsilon_1$ is the error tolerance. Here ${\eta }_{e}$$=$$5$, and $\epsilon_1$$=$$0$.

	Considering that the agent is hard to choose the `end' action because of huge searching space, the agent is forced to stop when it exceeds the maximum step allowed. This is described by the \textit{force ending reward}
	\begin{equation} \small \label{equ:Force Ending Reward }
	{{R}_{fe}}\left( {{E}_{t_m}} \right)=\left\{ \begin{matrix}
	+{{\eta }_{e}} & ~~~~\text{if} \left| {{E}_{t_m}} \right|\le\epsilon_1  \\
	-{{\eta }_{e}} & \text{otherwise}  \\
	\end{matrix} \right.\,,
	\end{equation}
	where ${E}_{t_m}$ is the absolute value error at the maximum step $t_m$.

	\paragraph{\textit{Intermediate Reward}.}
	In previous works~\cite{caicedo2015active,kong2017collaborative} that employ deep RL to deal with object localization,
	an intermediate reward is simply given according to the change of IoU.
	In counting, an optimal action can be computed to reach the balancing state faster.
	We thus introduce a \textit{guiding reward} to push the agent to choose the optimal action, defined by
	\begin{equation} \small \label{equ:Guiding Reward}
	{{R}_{g}}\left( {{E}_{t}},{{E}_{t-1}},{{{a}}_{t}},{{{a}}_{t}^g} \right)=\left\{ \begin{matrix}
	{{\eta }_{g}}  \\
	{{\eta }_{+ }}  \\
	{{\eta }_{- }}  \\
	\end{matrix} \right.\begin{matrix}
	&\text{if}\ {{a}_{t}}={{a}_{t}^g}  \\
	&~~~~~\text{if}\ {{E}_{t}}<{{E}_{t-1}}  \\
	&\text{otherwise}  \\
	\end{matrix}\,,
	\end{equation}
	where $a_t$ is the action chosen in the step $t$, and $a_t^g$ is the optimal action, given by
	\begin{equation} \small \label{equ:optimal}
	{{a}_{t}^g}=\underset{a}{\mathop{\arg \min }}\,\left| G\left(i\right)-\left( V^{i}_{t-1}+a \right) \right|\,.
	\end{equation}
	In our implementation, ${\eta}_{g}$$=$$+3$, ${\eta }_{+}$$=$$+1$, and ${\eta }_{-}$$=$$-1$.

	\algsetup{indent=2em}
	\begin{algorithm}[!t]
		\centering
		\renewcommand{\algorithmicrequire}{\textbf{Input:}}
		\renewcommand{\algorithmicensure}{\textbf{Output:}}
	    \caption{Training Procedure of LibraNet}
		\label{alg:overall}
		\begin{algorithmic}[1]\footnotesize
			\STATE Initialize a \textit{Buffer} $\leftarrow[~]$, the Q-network $N_Q^\theta$, and the backbone network $N_b$
			\FOR{ epoch $\leftarrow$ $0$ to NumEpochs  }
			\STATE Update the Q-network $N_Q^{\bar{\theta}}\leftarrow N_Q^{\theta}$
			\FORALL{ image $I$ in the training dataset }
			\STATE Compute the image feature $FV_I\leftarrow N_b\left(I\right)$
			\FORALL{patch $i$ in image $I$}
			\STATE Initialize $W_0^i\Leftarrow\{0,0,...\}$

			\STATE Fetch the ground-truth patch count $G\left( i \right)$
			\FOR{$t\leftarrow0$ to $T_e$}
			\STATE Obtain the state $s_{t}\leftarrow\{FV^i_I, W_{t}^i\}$
			\STATE Compute the Q value $Q_t\leftarrow N_Q^{\theta} \left(s_t\right)$
			\STATE Choose an action $a_t$ with $\epsilon$-greedy policy
			\STATE Compute the reward $r$ according to Sec.~\ref{subsec:Q learning}
			\STATE Update $W_{t+1}^i$ per Eq.~\eqref{equ: W update}
			\STATE Obtain the next state $s_{t+1}\leftarrow\{FV^i_I, W_{t+1}^i\}$
			\STATE \textit{Buffer} $\leftarrow(s_t,a_t,s_{t+1},r)$
			\ENDFOR
			\ENDFOR
			\STATE Sample a batch $B$ from the \textit{Buffer} to train $N_Q^{\theta}$ per Eq.~\eqref{equ:DDQN}
			\ENDFOR
			\ENDFOR
		\end{algorithmic}
	\end{algorithm}

	In our experiments, we find that, at the first several training epochs, the agent tends to choose large value operators that lead to overestimation. A possible explanation is that, because of the huge searching space, the agent cannot search for actions smoothly. To reach the balancing state faster, we propose a \textit{squeezing reward} to constrain the estimated value, defined by
	\begin{equation} \small\label{equ:Squeezed Reward}
	{{R}_{s}}=\left\{ \begin{matrix}
	{{R}_{g}}\left( {{E}_{t}},{{E}_{t-1}},{{a}_{t}},{{a}_{g}} \right)  & \text{if}~~S\left( {{V^i_t}},{{G\left(i\right)}} \right)=1  \\
	{{R}_{sg}}\left( {{E}_{t}},{{E}_{t-1}},{{a}_{t}},{{a}_{g}} \right) & \text{otherwise}  \\
	\end{matrix} \right.\,,
	\end{equation}
	where ${R}_{g}$ is the \textit{guiding reward} (Eq.~\eqref{equ:Guiding Reward}). $S\left({{V^i_t}},{G\left(i\right)} \right)$ decides whether $V^i_t$ is out of the tolerance range as
	\begin{equation} \small \label{equ: S}
	S\left( V^i_t,G\left(i\right) \right)=sign\left(G\left(i\right)\times \epsilon_2 - \left( V^i_t-G\left(i\right)\right) \right)\,,
	\end{equation}
	where $\epsilon_2$ is a tolerance range set to $0.5$ in this paper. If $S\left( V^i_t,G\left(i\right) \right)$$=$$-1$, we leverage a \textit{squeezed guiding reward} to squeeze the estimation within the tolerance range, defined by
	\begin{equation} \small \label{equ: Squeezed Guiding Reward}
	{{R}_{sg}}\left( {{E}_{t}},{{E}_{t-1}},{{{a}}_{t}},{{{a}}_{t}^g} \right)=\left\{ \begin{matrix}
	{{\eta }_{sg}}  \\
	{{\eta }_{s }}  \\
	\end{matrix} \right.\begin{matrix}
	& \text{if}~{{a}_{t}}={{a}_{t}^g}  \\
	& \text{otherwise}  \\
	\end{matrix}\,,
	\end{equation}
	where ${\eta}_{sg}$$=$$-1$, and ${\eta}_{s}$$=$$-3$. Notice that, in this reward function, all rewards are set to be negative such that the agent is encouraged to avoid choosing an action sequence that leads to overestimation.

	\paragraph{\textbf{Q Values (Needle)}.}
	In Q learning, the Q value of an action is an estimation of the accumulated reward after this action is taken, which takes the form
	\begin{equation} \small
	Q\left( s_t,a_t \right)=\left\{ \begin{matrix}
	r & \text{if}~a_t=\text{`end'}\ \text{or} \ t=t_m \\
	r+\gamma {{\max }_{a'}}Q\left( s_{t+1},a' \right) & \text{otherwise}  \\
	\end{matrix} \right.\,,
	\end{equation}
	where $r$ is the reward coming from either $R_e$, $R_{fe}$, $R_g$ or $R_{sg}$, the next state $s_{t+1}$ is acquired after the action $a_t$ is taken at the present state $s_t$, and $\gamma$ is the reward discount factor set to $0.9$ in our experiments. The Q value of each action is the output of DQN. It guides action selection and implies how the agent judges the scale balance. Hence Q value can be seen as the `needle' of the `counting scale'.

	\subsection{LibraNet}\label{subsec:Structure}
	Here we give an overview of LibraNet (Fig.~\ref{fig:overview}). LibraNet consists of two parts: a feature extraction backbone and a DQN. The backbone includes $5$ convolutional blocks of VGG16~\cite{simonyan2014very}. It aims to extract the feature map $FV_I$ of an image $I$. Each spatial feature vector $FV^i_I$ in $FV_I$ and its weight vector $W_t^i$ correspond to a $32\times32$ block in the original image. The backbone uses the model trained by~\cite{liu2019countingobject} and is then fixed when training the DQN.

	The core of LibraNet is the DQN. Its input is $FV^i_I$ and $W_t^i$. In each step of the training stage, $FV^i_I$ and $W_t^i$ are concatenated and sent to a two-layer multi-layer perception (MLP) with $1024$-dimensional hidden units in each layer, and the DQN outputs a $9$-dimensional Q value $Q_t$. An action $a_t$ chosen by $\epsilon$-greedy policy (Sec.~\ref{subsec:Implement Details}) is then concatenated with $W_t^i$ to obtain $W_{t+1}^i$ (Eq.~\eqref{equ: W update}). The estimation repeats until the `end' action is reached or exceeds $t_m$ steps. The output of DQN is the weighing vector $W_{T_e}^i$ for each patch $i$. When the weighing task terminates, $V^i_{T_e}$ is computed according to Eq.~\eqref{equ: V_T}.

	In the inferring stage, the agent chooses the action with the maximal Q value to obtain the weighing vector $W_{T_e}^i$ and the weighing value $V^i_{T_e}$ of each patch.
	Notice that $V^i_{T_e}$ is still the quantized count interval. It needs to be further mapped to a counting value with a class-count look-up table~\cite{liu2019countingobject}. Finally we can sum all patch counts to obtain the image count.

	\subsection{Implementation Details}\label{subsec:Implement Details}

	Following~\cite{mnih2015human}, we use a replay memory buffer~\cite{lin1993reinforcement} to remove correlations in the weighing process.
	We follow the standard DQN~\cite{mnih2015human} structure which has a Q-network and a target network. The target network computing the target Q value $({{\max }_{a'}}Q\left( s_{t+1},a' \right)$$+$$r)$ is fixed when training the Q-network,
	and we update the target network at the beginning of each epoch with the parameters of the Q-network. $\ell_1$ loss is used for optimization. The overall loss is defined by
	\begin{equation} \small \label{equ:DDQN}
	\ell={\sum\limits_{(s_t,a_t,s_{t+1},r)\in{U(B)}}\left| r+\gamma \underset{a'}{\mathop{\max }}\,N_Q^{\bar{\theta}}\left( s_{t+1},a'\right)-N_Q^{{\theta}}\left( s_t,a_t\right) \right|}/N\,,
	\end{equation}
	where $N_Q$ is LibraNet, $\theta$ and $\bar{\theta }$ are the parameters of the Q-network and the target network, respectively, $r$ is the reward, and $\gamma$ is the discount factor.

	During training,
	we follow the $\epsilon$-greedy policy: a random action is chosen either with a probability of $\epsilon$ or according to the maximum Q value. $\epsilon$ starts from $1$ and decreases
	to $0.1$ with a step of $0.05$.
	To reduce computation cost, we update the model when every $100$ samples are sent to the buffer. Considering that, the maximum quantized count interval is less than $80$, the maximum step $t_m$ is set to $8$ (the maximum value operation is $+10$). Algorithm~\ref{alg:overall} summarizes the training flow.
	We use SGD with a constant learning rate of $1e^{-5}$.

	Following~\cite{li2018csrnet},
	we crop $9$ $\frac{1}{2}$-resolution patches. These patches are mirrored to double the training set. For the UCF-QNRF dataset~\cite{idrees2018composition}, we follow BL~\cite{ma2019bayesian} to limit the shorter side of the image to be less than $2048$ pixels and to crop $512\times512$ patches for training.

	\section{Experimental Results}\label{sec:Results}
	Here we validate the effectiveness of LibraNet, visualize the weighing process, compare it against other state-of-the-art methods, demonstrate its cross-dataset generalization, justify each design choice, and show its generality as a plug-in. We report the mean absolute error (MAE) and (root) mean square error (MSE).

	\subsection{Visualization of the Weighing Process}\label{subsec:Visualization}
	\begin{figure}[!t]
		\centering
		\small
		\includegraphics[width=0.95\linewidth]{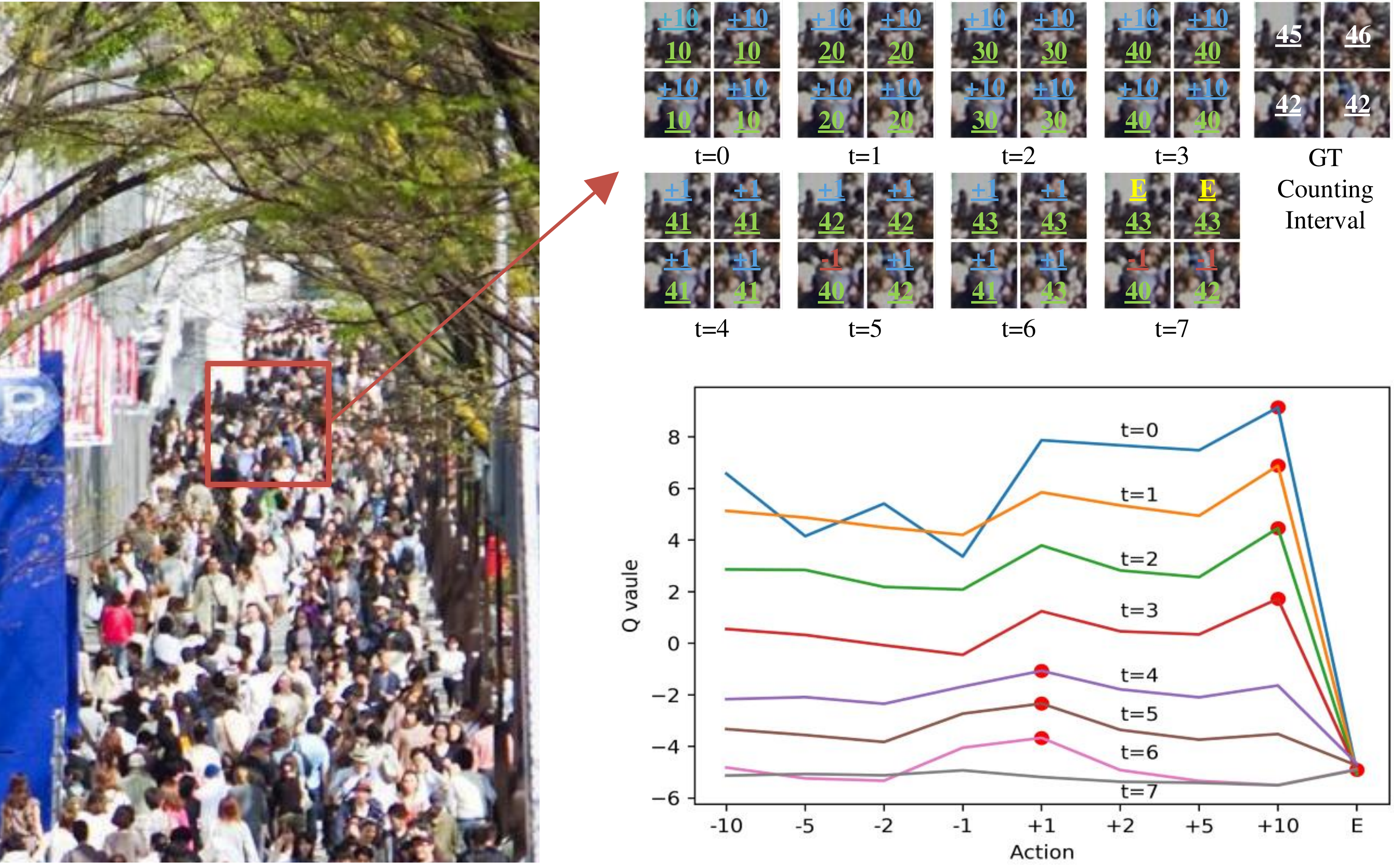}%
		\caption{Visualization of the inferring process of LibraNet. (upper right) Visualizations of action selection. We estimate the count interval for each $32\times32$ patch of the image. The weighing process is shown from $t$$=$$0$ to $t$$=$$7$, and the ground truth count intervals are shown in the right.
		For each patch, the lower green number is the accumulated value (the count interval), and the upper number is the value operator, including the value-increased operator (blue), the value-decreased operator (dull-red), and the ending operator `E' (yellow). (bottom right) Estimated Q values in each step of the upper left patch. The red point in each step is the Q value of the chosen action.
		}
		\label{fig:vision}
	\end{figure}

	To understand how LibraNet works, we visualize the inferring process of one sample in Fig.~\ref{fig:vision}. It can be seen that, in the first several steps, LibraNet tends to choose the action such that the estimation increases rapidly to approximate the ground truth. This is consistent with the target of \textit{guiding reward} (Eq.~\eqref{equ:Guiding Reward}). When the accumulated value is close to the ground truth, LibraNet begins to choose actions with small values. This is similar to how we weight a stuff using a scale. Once the accumulated value equals to the ground truth, the weighing process terminates. Notice that, even if the maximum step is reached, LibraNet still produces a relatively accurate estimation due to \textit{force ending reward} (Eq.~\eqref{equ:Force Ending Reward }).
	Interestingly, we find that the agent chooses positive actions more frequently than negative ones, because i) the initial value is $0$, and the target count is either $0$ or positive. Thus, the agent tends to choose positive actions to approximate the ground truth, and ii) we design a squeeze guide reward (Eq.~\eqref{equ: Squeezed Guiding Reward}) to avoid overestimation. This reward penalizes overestimation and further decreases the frequency of selecting negative actions.

	To further analyze why the agent chooses certain actions, we visualize Q values of the top left patch. The ground truth count interval is $45$, and the agent chooses four consecutive $+10$, three $+1$ and one $End$ actions. The final estimated interval is $43$. In the first $4$ steps, Q values excluding $End$ are greater than $0$ and have a clear distinction. It means that the agent is confident with its action selection. After four steps, the accumulated value is $40$, which closes to the ground truth. In the last $4$ step, Q values are less than $0$, and the differences between each action is small, which implies the agent is aware of the closeness to the ground truth. To avoid overestimation, the agent becomes cautious to avoid a significantly wrong decision. Even if the final weighing value does not strictly equal to the ground truth, the estimation is not likely to shift away from the ground truth significantly.
	We can see that LibraNet follows exactly how a scale weighs a stuff, which means LibraNet indeed learns what we expect it to learn.

	\subsection{Comparison with State of the Art}\label{subsec:STOA}
	We evaluate our method on three public crowd counting benchmarks: ShanghaiTech, UCF\_CC\_50 and UCF-QNRF.

	\begin{table}[!t] \scriptsize
		\centering
		\renewcommand\arraystretch{1.05}
		\caption{Comparison with state-of-the-art approaches on three crowd counting benchmarks. The lowest errors are boldfaced}
		\label{tab:STOA}
		\setlength{\tabcolsep}{2.7mm}{
			\begin{tabular}{ r c c c c c c c c }\hline
				\multirow{2}[1]*{Method}& \multicolumn{2}{c}{SHT Part\_A} & \multicolumn{2}{c }{SHT Part\_B} &\multicolumn{2}{c}{UCF\_QNRF}&\multicolumn{2}{c}{UCF\_CC\_50}\\
				& MAE & MSE & MAE & MSE & MAE & MSE & MAE & MSE\\\hline
				DRSAN~\cite{liuijcai2018crowd}
				&69.3& 96.4&11.1&18.2&---&---&219.2&250.2\\
				CSRNet~\cite{li2018csrnet}
				&68.2&115.0&10.6&16.0&---&---&266.1&397.5\\
				TEDnet~\cite{jiang2019crowd}
				&64.2&109.1&8.2&12.8&113&188&249.4&354.5\\
				SPN+L2SM~\cite{xu2019learn}
				&64.2&98.4 &7.2&11.1&104.7&173.6&188.4&315.3\\
				BCNet~\cite{liu2019countingobject}
				&62.8&102.0&8.6&16.4&118  &192&239.6&322.2  \\
				BL~\cite{ma2019bayesian}
				&62.8&101.8&7.7&12.7&88.7&154.8&229.3&308.2\\

				CAN~\cite{liu2019context}
				&62.3&100.0&7.8&12.2&107&183&212.2&\textbf{243.7} \\

				MBTTBF~\cite{sindagi2019multi}
				&60.2&94.1 &8.0&15.5&97.5&165.2&233.1&300.9\\
				PGCNet~\cite{yan2019perspective}
				&57.0& \textbf{86.0}&8.8&13.7&---&---&244.6&361.2\\
				S-DCNet~\cite{Xiong2019CLOSE}
				&58.3& 95.0&\textbf{6.7}&\textbf{10.7}&104.4&176.1 &204.2&301.3\\\hline

				LibraNet &\textbf{55.9}& 97.1&7.3&11.3&\textbf{88.1}&\textbf{143.7}&\textbf{181.2}&262.2\\\hline
			\end{tabular}
		}
	\end{table}

	The ShanghaiTech (SHT) Dataset~\cite{zhang2016single} includes $1,198$ crowd images with $330,165$ head annotations. It has two parts: part A includes $482$ images with varying resolution collected from Internet; part B includes $716$ images of the same resolution collected from street surveillance videos. In part A, $300$ images are used for training, and other $182$ images for testing. Part B adopts $400$ images for training and $316$ images for testing.
	Results are shown in Table~\ref{tab:STOA}.  We compare our method against other $10$ state-of-the-art methods and report the best MAE in part A and comparable performance on part B.

	The UCF\_CC\_50 Dataset~\cite{Idrees2013Multi} is a challenging crowd counting dataset with only $50$ images. By contrast, there are $63,705$ people annotations, so the scenes are extremely congested.
	We employ $5$-fold cross-validation when reporting the results and also compare LibraNet with other state-of-the-art approaches.
	The results shown in Table~\ref{tab:STOA} verify that LibraNet outperforms other competitors and reports the best performance in MAE.

	The UCF-QNRF Dataset~\cite{idrees2018composition} is a recent %
	high-solution crowd counting dataset, which includes $1,535$ images with $1,251,642$ annotations. The images are officially split into two parts: $1201$ images for training and $334$ for testing.
	We compare LibraNet with $7$ recent methods. The results in Table~\ref{tab:STOA} illustrate our method outperforms state-of-the-art methods in both MAE and MSE.

	\subsection{Cross-Dataset Generalization}\label{subsec:Cross data}

	To demonstrate the generalization of LibraNet, we conduct cross-dataset experiments by training the model on one dataset but testing on the other one.
	Results are shown in Table~\ref{tab:Cross experiments}.
	LibraNet shows consistently better generalization performance than other competitors across all transfer settings.

	\begin{table}[!t] \scriptsize
		\centering
		\renewcommand\arraystretch{1.05}
		\caption{Cross-dataset evaluations on the ShanghaiTech (A and B) and UCF-QNRF (QNRF) datasets}
		\label{tab:Cross experiments}
		\setlength{\tabcolsep}{0.8mm} {
			\begin{tabular}{r c c c c c c c c c c c c }
				\hline
				\multirow{2}[1]*{Method}& \multicolumn{2}{c}{A $\rightarrow$ B}& \multicolumn{2}{c}{A$\rightarrow$QNRF}& \multicolumn{2}{c}{B$\rightarrow$A }& \multicolumn{2}{c}{B$\rightarrow$QNRF}& \multicolumn{2}{c}{QNRF$\rightarrow$A }& \multicolumn{2}{c}{QNRF$\rightarrow$B}\\

				& MAE& MSE & MAE & MSE & MAE  & MSE  & MAE  & MSE & MAE  & MSE & MAE  & MSE   \\\hline
				MCNN~\cite{zhang2016single}  &85.2&142.3&---  &---  &221.4 &357.8 &--- &---&--- &---&--- &---\\
				D-ConvNet~\cite{shi2018crowd}                 &49.1& 99.2&---  &---  &140.4 &226.1 &--- &---&--- &---&--- &---\\
				SPN+L2SM~\cite{xu2019learn}  &21.2&38.7 &227.2&405.2&126.8&203.9&---&---                   &73.4&119.4&---&---\\
				BCNet~\cite{liu2019countingobject}&20.5&37.9&131.9&230.6&138.6&230.0&240.0&419.6&71.3&123.7&16.1&26.1\\
				BL~\cite{ma2019bayesian}     & ---&---  &---  &---&---  &---&---&---                                         &69.8&123.8          &15.3&26.5\\\hline
				LibraNet                   &\textbf{11.9}&\textbf{20.7}&\textbf{127.9}&\textbf{204.9}&\textbf{98.3}&\textbf{167.9}                   &\textbf{224.2}&\textbf{405.3}              &\textbf{67.0}&\textbf{109.2}  &\textbf{11.9}&\textbf{22.0}\\\hline
		\end{tabular}}

    	\centering
    	\caption{Ablation study on the ShanghaiTech Part A dataset}
    	\label{tab:Ablation study}
    	\renewcommand\arraystretch{1.05}
    	\setlength{\tabcolsep}{22pt}{
    		\begin{tabular}{ r c c}\hline
    			\small
    			Method            & MAE   & MSE\\\hline
    			BCNet~\cite{liu2019countingobject}  & 62.8 & 102.0\\
    			Imitation Learning~\cite{hussein2017imitation} & 64.7 & 102.8\\\hline
    			W/O Guiding              & 149.8& 261.3\\
    			W/O Force Ending         & 62.7 & 104.3\\
    			W/O Squeezing            & 63.5 & 102.7\\
    			Full Designs             & \textbf{55.9} & \textbf{97.1}\\\hline
    		\end{tabular}
    	}

    	\caption{GAME on the ShanghaiTech Part A dataset}
    	\label{tab:game}
    	\centering
    	\renewcommand\arraystretch{1.05}
    	\setlength{\tabcolsep}{10pt}
    	{
    		\begin{tabular}{r c c c c }\hline
    			\small
    			& GAME0  & GAME1  & GAME2  & GAME3  \\\hline
    			BCNet~\cite{liu2019countingobject}&62.8&73.3&87.0&116.7\\
    			LibraNet& \textbf{55.9}&\textbf{68.0}& \textbf{82.1}&\textbf{113.1}\\\hline

    		\end{tabular}
    	}
	\end{table}

	\subsection{Ablation Study}\label{subsec:Ablation study}
    Here we validate basic design choices of LibraNet on the SHT Part\_A dataset~\cite{zhang2016single}. The results are shown in Table~\ref{tab:Ablation study}.

	\paragraph{\textbf{Local Accuracy}.}
	BCNet is the blockwise classification network proposed by~\cite{liu2019countingobject}. This is our direct baseline, because LibraNet uses the backbone pretrained by~\cite{liu2019countingobject}.
	Besides the image-level error, we also report the Grid Average Mean absolute Error (GAME)~\cite{Guerrero2015Extremely} in Table~\ref{tab:game}. GAME assesses patch-level counting accuracy.
	LibraNet outperforms BCNet in all GAME metrics, which suggests that LibraNet generates more locally accurate patch counts than BCNet. We believe this may be the reason why LibraNet significantly reduces the image-level error.

	\paragraph{\textbf{Optimal Action}.}
	In Sec.~\ref{subsec:Q learning}, we compute the optimal action to reach the balancing state faster. \textit{Is it sufficient to learn a weighing model that only chooses the optimal action?}
	To justify this, we build another baseline `Imitation Learning'~\cite{hussein2017imitation} with the following optimization target

	\begin{equation}\label{equ:  Imitation Learning}
	\underset{\theta} {\mathop{\max }} \sum_{i \in I}\sum_{t=0}^{T_e} \sum_{a=0}^{A_N} [a=a_{i,t}^g] \log\left( N^{\theta}_M \left(i,W^i_t,a\right)  \right) \,,
	\end{equation}
	where $a_{i,t}^g$ is the optimal action (Eq.~\eqref{equ:optimal}) of time $t$ in patch $i$, $A_N$ is the number of pre-defined action, $N^{\theta}_M$ is a sequential decision network,
	and $N^{\theta}_M \left(i,W^i_t,a\right)$ computes the probability of $a$-th action in $i$-th patch. In each step, $N^{\theta}_M$ selects the action with the maximum probability. Results in Table~\ref{tab:Ablation study} show that \textit{learning with only the optimal action is insufficient}.

	\begin{table}[!t] \scriptsize
		\caption{Sensitivity analysis of the maximum step on the ShanghaiTech Part A dataset}
		\label{tab:step}
		\centering
		\renewcommand\arraystretch{1.05}
		\setlength{\tabcolsep}{13.25pt}
		{
			\begin{tabular}{c c c c c c c c }\hline
				\small
				Step& 4  & 6  & 8  & 10 & 12 & 14 &16 \\\hline
				MAE& 126.3&59.0& 55.9&57.7& 62.5 & 60.7 &56.9\\
				MSE& 243.0&106.6&97.1& 99.2&101.3 &100.5&93.9\\\hline

			\end{tabular}
		}
		\caption{Sensitivity analysis of the tolerance range on the ShanghaiTech Part A dataset}
		\label{tab:tolerance}
		\centering
		\renewcommand\arraystretch{1.05}
		\setlength{\tabcolsep}{19.5pt}
		{
			\begin{tabular}{c c c c c c }\hline
				\small
				Range& 0.1  & 0.3  & 0.5  & 0.7&0.9  \\\hline
				MAE& 61.0&59.7& \textbf{55.9}&60.1& 59.9\\
				MSE&104.5&100.1&97.1& \textbf{96.8}&103.3\\\hline

			\end{tabular}
		}
		\centering
		\caption{LibraNet as a plug-in}
		\label{tab:Adapt}
		\renewcommand\arraystretch{1.05}
		\setlength{\tabcolsep}{12pt}{
			\begin{tabular}{l c c}\hline

				\small
				Method            & MAE   & MSE\\\hline
				ImageNet Regression  & 156.2& 259.9\\
				ImageNet Classification  & 140.4& 230.3\\
				ImageNet Regression+LibraNet        & 126.6& 211.1\\
				ImageNet Classification+LibraNet        & 119.7& 203.4\\  \hline
				TasselNetv2$^{\dag}$~\cite{xiong2019tasselnetv2}      & 68.6& 110.2\\
				TasselNetv2$^{\dag}$+LibraNet                         & 64.7& 100.6\\\hline
				Blockwise Classification~\cite{liu2019countingobject} & 62.8 & 102.0\\
				Blockwise Classification+LibraNet                     & \textbf{55.9} & \textbf{97.1}\\\hline
			\end{tabular}
		}
	\end{table}

	\paragraph{\textbf{Designed Rewards}.} From the $3$-th to the $5$-th rows of Table~\ref{tab:Ablation study}, we present the ablative studies on modified rewards. `W/O~Guiding' means training LibraNet without the `guiding reward' (Eq.~\eqref{equ:Guiding Reward}) which simply sets $+1$ to error-decreased action and $-1$ to error-increased action, `W/O~Force~Ending' means training LibraNet without the `force ending reward' (Eq.~\eqref{equ:Force Ending Reward }), and `W/O~Squeezing' means training LibraNet without the `squeezing reward' (Eq.~\eqref{equ: Squeezed Guiding Reward}). It is clear that all designed rewards benefit counting.

	\paragraph{\textbf{Parameters Sensitivity}.} To analyze the impact of the maximum action step $t_m$, we train LibraNet with
	$t_m$ ranging from $4$ to $16$ on the SHT Part A dataset. Results are shown in Table~\ref{tab:step}. When $t_m$ is not sufficient, LibraNet works poorly, because LibraNet cannot reach the neighborhood of ground truth even if the maximum value operation can be chosen in each step.
	We set $t_m=8$ in all other experiments. We also evaluate the effect of the tolerance range ($\epsilon_2$) in Eq.~\eqref{equ:Squeezed Reward}. Results are shown in Table~\ref{tab:tolerance}. We observe that, LibraNet is not sensitive to this parameter, and the best result is achieved when $\epsilon_2=0.5$ on the SHT Part A dataset. We thus fix $\epsilon_2=0.5$. Furthermore, we analyze the effect of randomness. Following~\cite{van2016deep}, we run LibraNet for $6$ times on the SHT Part A with different random seeds. The MAE is $56.4\pm1.8$, and MSE is $97.8\pm2.3$, which suggests LibraNet is not sensitive to randomness.

    \paragraph{\textbf{Execution Speed}.} Finally, we report the %
    speed of LibraNet on a platform with RTX $2060$ $6$ GB GPU and Intel i7-9750H CPU. It takes $158$ ms to process an $1080\times720$ image, including $142$ ms on backbone and $16$ms on LibraNet. The result illustrates that LibraNet only introduces
    negligible
    computation costs.

    \subsection{LibraNet as a Plug-in}\label{subsec:Adapt}

    To show that LibraNet is a general idea and the pretraining with~\cite{liu2019countingobject} is not the only opinion, here we apply LibraNet as a plug-in to other counting/pretrained models. Results are shown in Table~\ref{tab:Adapt}.

	First we attach LibraNet to a regression baseline and a classification baseline with ImageNet-pretrained VGG16~\cite{simonyan2014very}.
	The VGG16 is fixed and concatenated with a trainable $1\times1\times C$ or a $1\times1\times1$ convolution kernel to classify counting intervals or to regress patch counts.
	By using LibraNet, we observe more than $10\%$ relative improvements over the regression and classification baselines.
	In addition, it can be observed that `ImageNet Regression/Classifiaction+LibraNet' exhibits significantly worse performance than other comparing approaches.
	This suggests that pretraining the feature extraction backbone is important for counting. Such results are consistent with a recent observation on visual question answering systems~\cite{chattopadhyay2017counting} that \textit{CNN features contain little information relevant to counting}~\cite{stahl2018divide}.

    The second model is a regression-based blockwise counter---TasselNetv2$^{\dag}$~\cite{xiong2019tasselnetv2}. `TasselNetv2$^{\dag}$+LibraNet' means extracting the feature map by the backbone pretrained by TasselNetv2$^{\dag}$ and then sending them to DQN to estimate the count. To adapt to regression-based weighing where the count values is continuous,
    we modify the pre-defined action pool $a=\{$ $-5$, $-2$, $-1$, $-0.5$, $-0.2$, $-0.1$, $-0.05$, $-0.02$, $-0.01$, $0.01$, $0.02$, $0.05$, $0.1$, $0.2$, $0.5$, $1$, $2$, $5$ $\}$.
	Results show that `TasselNetv2$^{\dag}$+LibraNet' outperforms TasselNetv2, which illustrates the idea of scale weighing is also effective for the regression-based counter.

	\section{Conclusion}\label{sec:Conlusion}

	In this work, we have introduced a novel sequential decision paradigm to
	tackle
	crowd counting, which is inspired by the behavior of human counting and scale weighing. We implement scale weighing %
	using
	deep RL and present a new counting model LibraNet. Experiments verify the effectiveness of LibraNet and explain how it works. For future work, we plan to extend LibraNet to other regression tasks.
	We believe that
	scale weighing is a general idea that may not be
	limited to counting.

\section*{Appendix}

\appendix

\section{Discretization and Inverse-Quantization  }
In this section, we illustrate the generation of counting intervals (`Discretization') and
inverse-quantization~\cite{liu2019countingobject} in detail.

\subsection{Counting Intervals Generation}
First, given a map of dotted annotations, it is convoluted by the Gaussian kernel to compute the density map $\mathcal{D}\left(p \right) $~\cite{lempitsky2010learning}, which takes the form
\begin{equation}\label{equ:density map}
\mathcal{D}\left(p \right) = \sum_{i=1}^N \delta \left(p - D_i \right) \ast G_{\sigma_i} \left(p\right)
\end{equation}
where $p \in I$ is a pixel in the image $I$, $D_i$ is the $i$-th dot annotation of $I$, and $G_{\sigma_i}$ is a Gaussian kernel with the variance of $\sigma_i$. In this paper, we employ the adaptive Gaussian kernel~\cite{zhang2016single}, whose variance is defined by
\begin{equation}\label{equ:adaptive}
\sigma_i  = \beta \bar{d}_i \,,
\end{equation}
where $\bar{d}_i$ is the average distance between the dot point ${D_i}$ and its $3$-nearest dot points, $\beta$ is a hyperparameter which is set to $0.3$ following~\cite{zhang2016single}.

Further, the density map is summed at patch-level to compute patch count $\mathcal{N} $~\cite{lu2017tasselnet}
\begin{equation}\label{equ:count map}
\mathcal{N}_i  =\sum_{p_b \in P_i} \mathcal{D}\left( p_b\right) \,,
\end{equation}
where $P_i$ is the $i$-th patch of the image $I$. %

Finally, the count value is quantized to compute the count interval $\mathcal{C}$~\cite{liu2019countingobject}
\begin{equation}\label{equ:class map}
\mathcal{C}_i= {\mathcal{Q}}{ \left( {\mathcal{N}_i} \right) }
={ \left\{ {\begin{array}{*{20}{l}}
			{0}&{\mathcal{N}_i=0}\\
			{max{ \left( {floor{ \left( {\frac{{log{ \left( {\mathcal{N}_i} \right) }-l}}{{w}}+2} \right) },1} \right) }}&{Otherwise}
			\end{array}}\right., }
\end{equation}
where $w$ is the width of the quantized interval in log space, $l$ is a hyperparameter which means the interval of $\left(0, e^l\right)$ is divided as an independent class~\cite{liu2019countingobject}. In this paper, we set $w=0.1$ and $l=-2$.

\subsection{Inverse-Quantization}
During testing, the counting value is recovered from the counting interval by the inverse-quantization $\mathcal{IQ}$~\cite{liu2019countingobject}, i.e.,
\begin{equation}\label{equ:inver quantization }
{\mathcal{N}_i= {\mathcal{IQ}}{ \left( {\mathcal{C}_i} \right) }
={ \left\{ {\begin{array}{*{20}{l}}
{0}&{\mathcal{C}_i=0}\\
{\frac{{1}}{{2}}exp{ \left( {l+w{ \left( {\mathcal{C}_i-1} \right) }} \right) }}&{\mathcal{C}_i=1}\\
{\frac{{1}}{{2}}exp{ \left( {l+w{ \left( {\mathcal{C}_i-2} \right) }} \right) }+\frac{{1}}{{2}}exp{ \left( {l+w{ \left( {\mathcal{C}_i-1} \right) }} \right) }}&{Otherwise}
\end{array}}\right.}}
\end{equation}

\section{More Qualitative Results}

\begin{figure}[h]
	\centering
	\includegraphics[width=\linewidth]{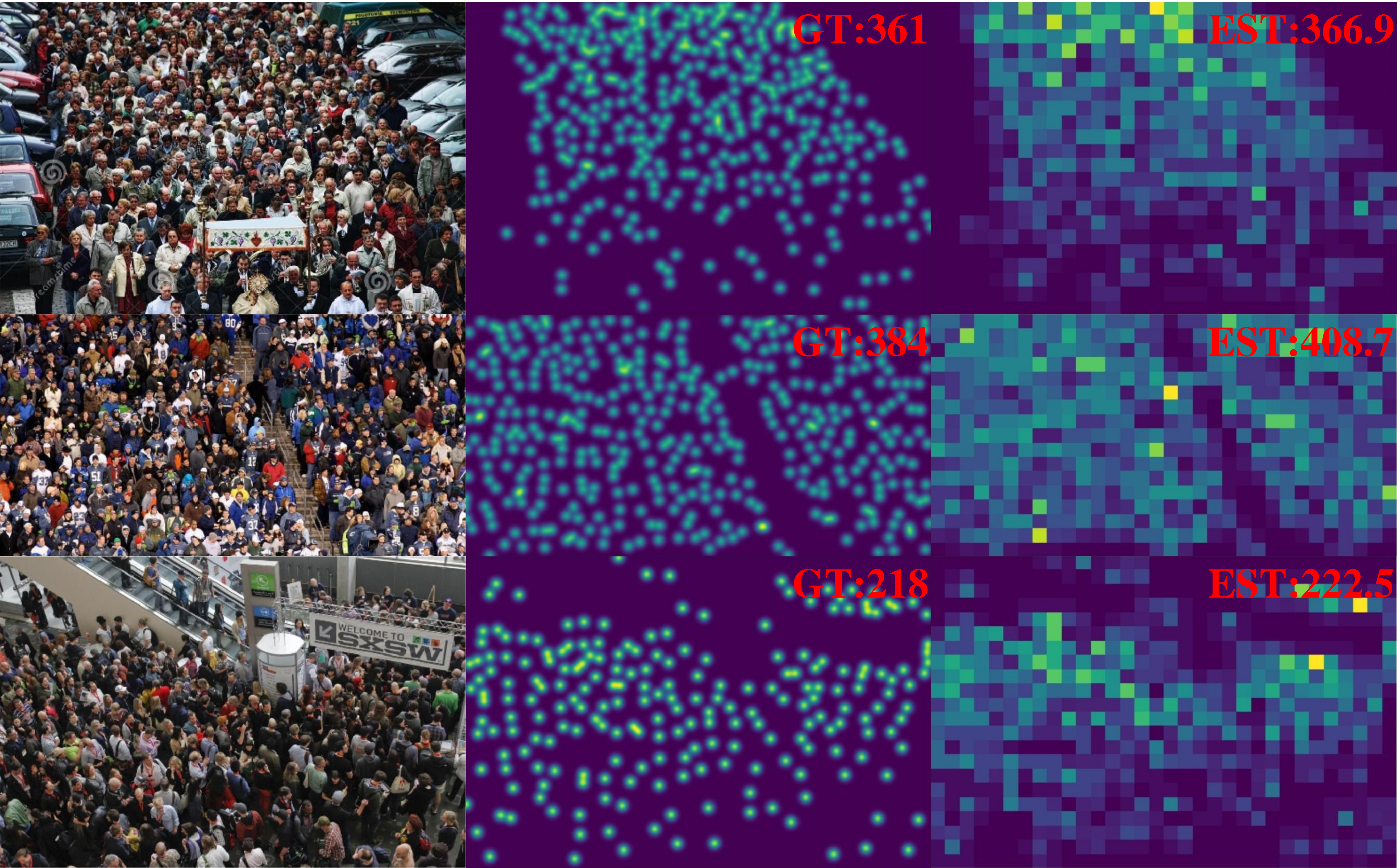}\\
	\caption{ Qualitative results of LibraNet on the ShanghaiTech part\_A dataset. From left to right, there are testing images, density maps with ground-truth counts, and our estimated results.}
	\label{fig:A}
\end{figure}

\begin{figure}[h]
	\centering
	\includegraphics[width=\linewidth]{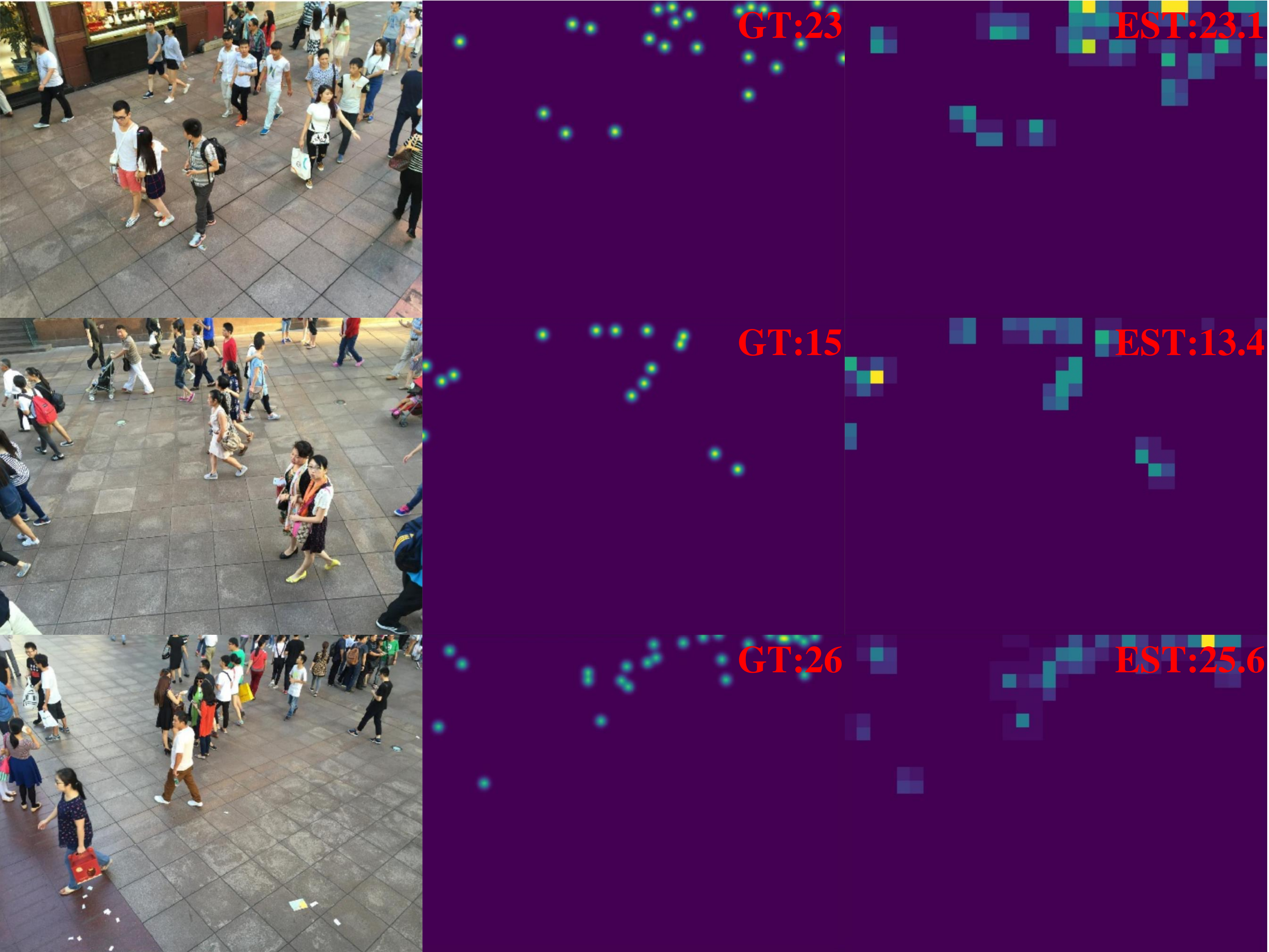}\\
	\caption{ Qualitative results of LibraNet on the ShanghaiTech part\_B dataset. From left to right, there are testing images, density maps with ground-truth counts, and our estimated results.}
	\label{fig:B}
\end{figure}

\begin{figure*}[h]
	\centering
	\includegraphics[width=\linewidth]{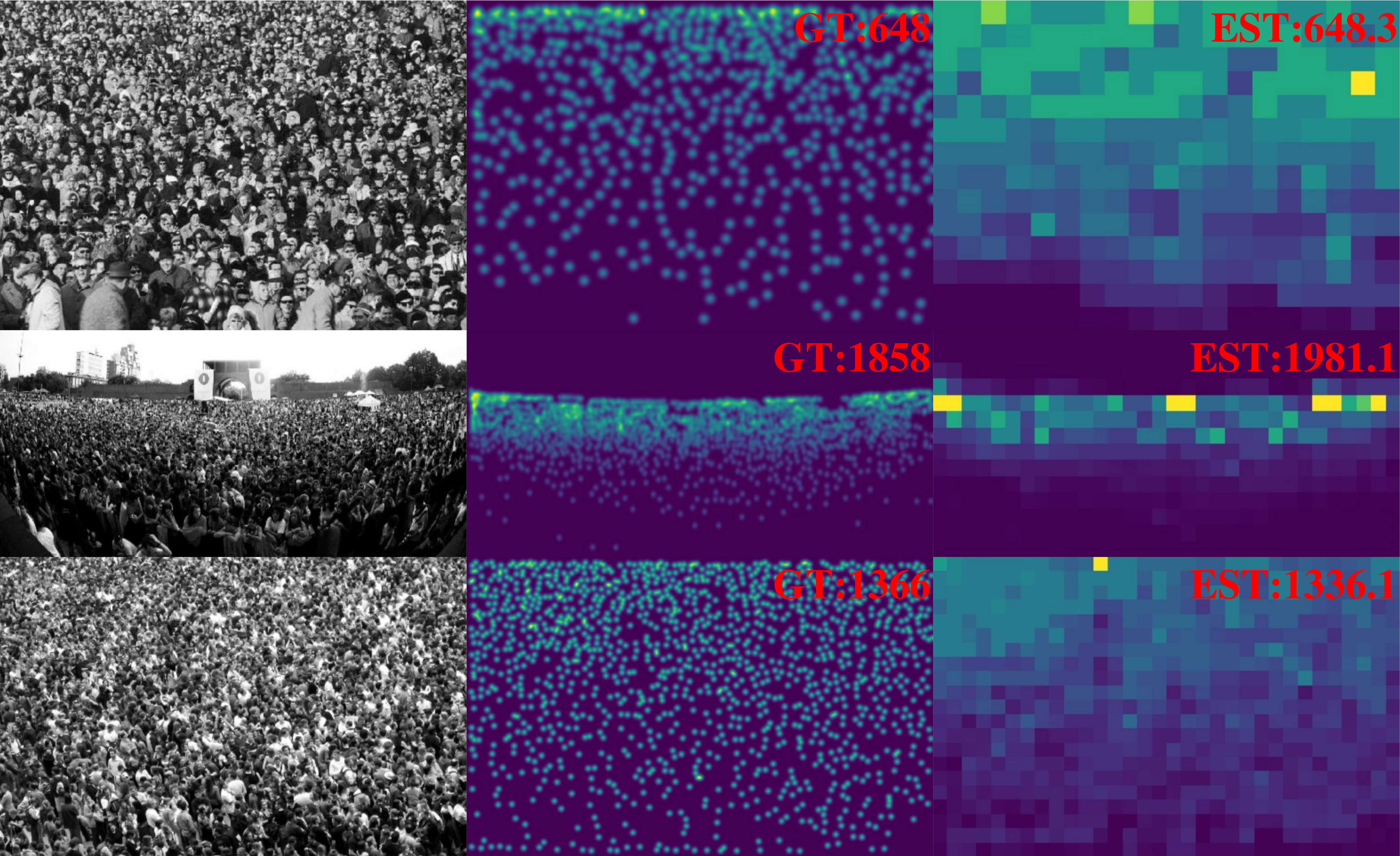}
	\caption{Qualitative results of LibraNet on the UCF\_CC\_50 dataset. From left to right, there are testing images, density maps with ground-truth counts, and our estimated results.}
	\label{fig:50}
\end{figure*}

\begin{figure*}[h]
	\centering
	\includegraphics[width=\linewidth]{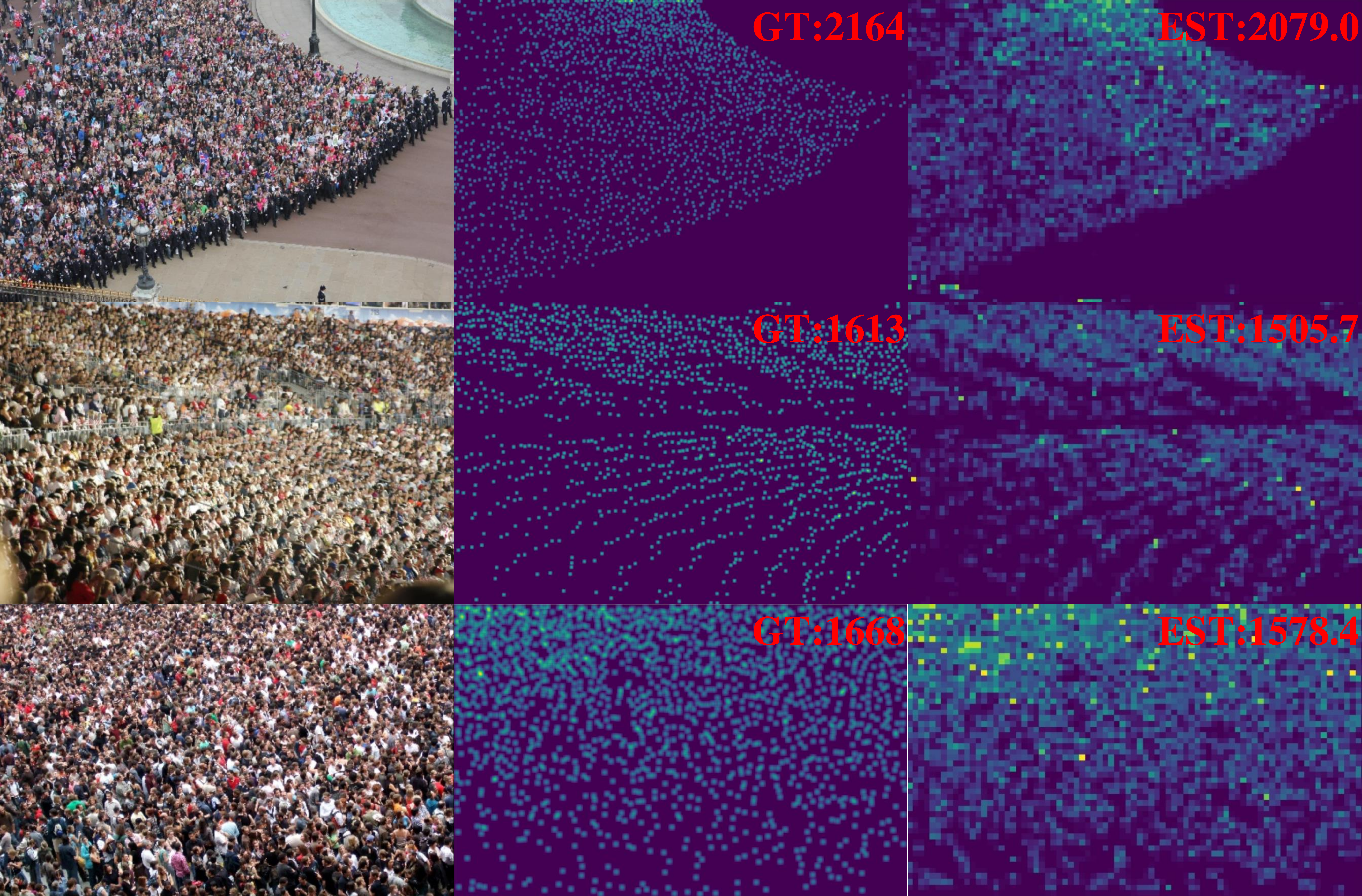}
	\caption{Qualitative results of LibraNet on the UCF-QNRF dataset. From left to right, there are testing images, density maps with ground-truth counts, and our estimated results.}
	\label{fig:QNRF}
\end{figure*}

\begin{figure*}[h]
	\centering
	\includegraphics[width=0.925\linewidth]{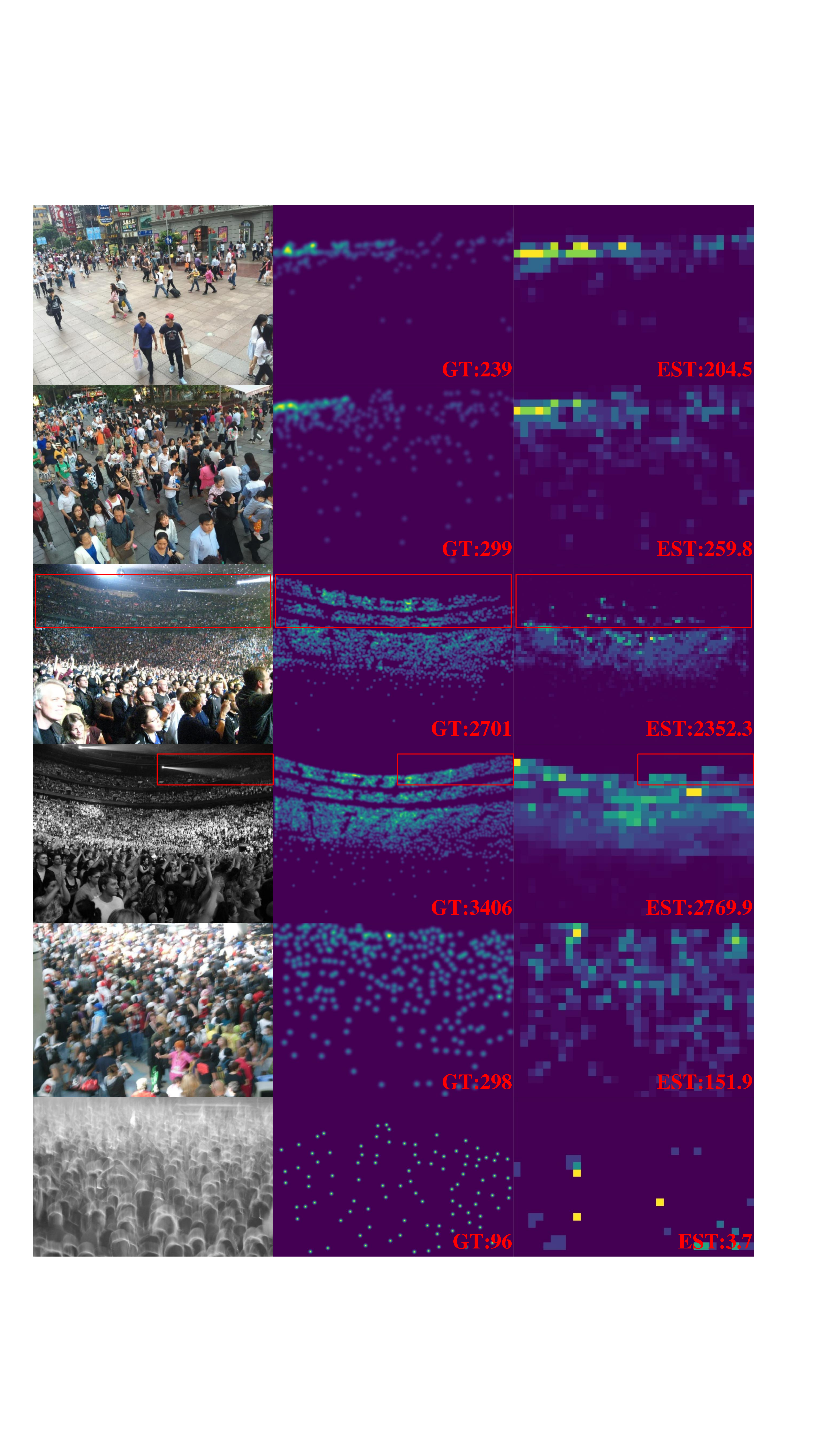}
	\caption{Failure cases. The first $2$ cases from the ShanghaiTech Part\_B dataset show that our method suffers from the training bias caused by long-tailed distribution, leading to under-estimations. The $3$-rd and $4$-th rows demonstrate that LibraNet can be affected by illumination variations. The last $2$ rows illustrate that our method fails due to blurry appearance. In each row, from left to right, there are testing images, density maps with ground-truth counts, and our estimated results.}
	\label{fig:failurecase}
\end{figure*}

\clearpage

	\paragraph{\textbf{Acknowledgement}.}
	This work is supported by the Natural Science Foundation of China under Grant No.\  61876211 and Grant No.\  U1913602.

	\bibliographystyle{splncs04}
	\bibliography{egbib}
\end{document}